\documentclass[conference]{IEEEtran}
\IEEEoverridecommandlockouts

\usepackage{booktabs,multirow}
\usepackage{amsmath,amssymb,amsfonts}
\usepackage{graphicx}
\usepackage{orcidlink}
\usepackage{hyperref}	
\hypersetup{colorlinks=true, citecolor=black, linkcolor=black, urlcolor=black}

\begin{document}

\title{Detecting Pen-In-Air States from Video: A Proof-of-Concept Toward Complementary Handwriting Analysis
\thanks{This work was funded by IMT Mines Ales and the Occitanie Region, France, through the AVIAREPTE doctoral project, as part of the Emergence 2025 program.}
}

\author{
    \IEEEauthorblockN{
    Lauren Sismeiro\IEEEauthorrefmark{1}\orcidlink{0009-0005-0898-5152},
    Rémy Plastre\IEEEauthorrefmark{4}\orcidlink{0009-0008-5522-5369},
    Binbin Xu\IEEEauthorrefmark{1}\orcidlink{0000-0002-0822-5250},
    Frédéric Puyjarinet\IEEEauthorrefmark{1}\orcidlink{0000-0003-3788-1473},
    Gérard Dray\IEEEauthorrefmark{1}\orcidlink{0000-0003-1525-5682}
    }
    \IEEEauthorblockA{\IEEEauthorrefmark{1}EuroMov Digital Health in Motion, Univ Montpellier, IMT Mines Ales, France}
    \IEEEauthorblockA{\IEEEauthorrefmark{4}IMT Mines Ales, France}
    \IEEEauthorblockA{
    \small
    \texttt{lauren.sismeiro@mines-ales.fr},
    \texttt{remy.plastre@mines-ales.org}
    }
}

\maketitle

\begin{abstract}
Dynamic aspects of handwriting are critical for assessing developmental disorders such as dysgraphia and are typically captured using digitizing tablets. However, tablet-based sensing restricts analysis of Pen-Up behavior to a short proximity range above the writing surface, potentially missing high-lift in-air movements. 
As a proof of concept, we investigate whether top-view video can provide a complementary source of information for inferring pen-contact states without relying on tablet proximity sensing. We propose an interpretable hybrid pipeline combining pen-tip tracking using a YOLO-based detector with kinematic feature extraction and machine learning classification. A pilot dataset of diverse handwriting videos was manually annotated at the frame level and evaluation used a Leave-One-Video-Out (LOVO) protocol. The method achieved reliable event-level detection of Pen-Up segments, with an $F_2$ score up to $0.805$, consistent with the emphasis on recall in a screening-oriented setting.
These results support the feasibility of video-based Pen-Up detection as a low-cost and non-intrusive complement to digitizing tablets, and provide a foundation for future large-scale studies.
\end{abstract}




\section{Introduction}

Handwriting is a complex perceptual-motor activity involving the integration of cognitive, linguistic, and fine motor processes. In clinical practice, the diagnosis of developmental dysgraphia, a specific learning disorder affecting handwriting quality and speed, primarily relies on the BHK (Brave Handwriting Kinder) scale \cite{charles_bhk_2004}. Targeted at children in elementary school (grades 1 to 5), the test involves a five-minute copying task evaluated through thirteen manual items. However, the reliability of BHK scoring remains debated in the literature \cite{murnani_handwriting_2025}. As the evaluation is based on the visual inspection of a written trace, it is partly subjective and can lead to variability between raters. Moreover, the BHK scale relies on post-hoc evaluation of the static written trace and does not capture the kinematic dynamics of handwriting, which are known to provide valuable information about motor control and planning.

To overcome these limitations, digitizing tablets have been used to record handwriting kinematics, enabling the extraction of dynamic features such as velocity, acceleration, and pressure. Combined with machine learning techniques, these recordings can help distinguish dysgraphic individuals from those with typical handwriting development \cite{asselborn_automated_2018, drotar_dysgraphia_2020, kunhoth_exploration_2023}. Among these features, in-air movements, corresponding to pen trajectories above the writing surface, have been shown to be particularly informative for distinguishing writing profiles \cite{rosenblum_air_2003}. In particular, recent studies have highlighted that on-surface features alone may be insufficient for reliable classification, whereas combining on-surface and in-air attributes significantly improves classification performance \cite{kunhoth_exploration_2023}.

However, a major limitation of digitizing tablets is their inability to capture pen motion beyond a short sensing range, typically a few millimeters above the surface \cite{gilhodes_chapitre_2023}. As a result, high-lift movements remain unobserved, leading to a loss of potentially informative information. This sensing gap prevents the analysis of high-lift pauses, which may carry information relevant to handwriting assessment, especially in case of dysgraphia.

Video-based approaches have previously been explored for handwriting analysis, mainly for recognition tasks \cite{fink_video-based_2001, munich_visual_2002}. However, their potential as a complementary modality for inferring pen-contact states and characterizing in-air movements remains underexplored.

To address this limitation, we investigate in this study the use of top-view video as a complementary modality to infer pen-contact states alongside digitizing tablets. Rather than aiming for a fully optimized system, we adopt a proof-of-concept perspective, focusing on the feasibility of extracting meaningful kinematic information from standard video recordings. We propose a modular pipeline combining pen-tip tracking and kinematic modeling, with an emphasis on simplicity and interpretability.

Given the limited dataset, this study should be regarded as a preliminary investigation rather than a demonstration of broad generalizability. Its objective is to evaluate whether a video-based approach can reliably detect pen-contact state transitions, particularly Pen-Up events, and to identify key methodological components for future large-scale developments.


\section{Materials and Methods}

\begin{figure*}
\centering
\includegraphics[width=0.95\textwidth]{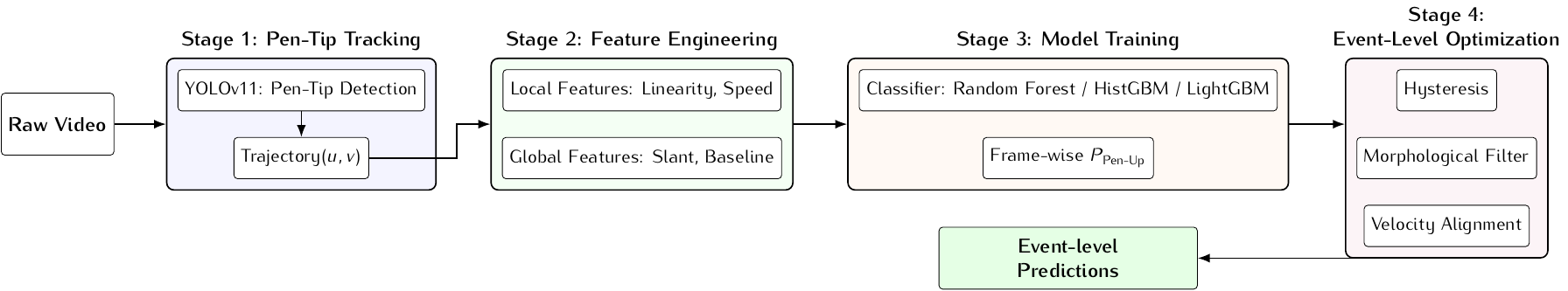}
\caption{Overview of the proposed hybrid pipeline for pen-contact state detection from top-view handwriting videos.}
\label{fig:pipeline}
\end{figure*}


\subsection{Data Collection and Annotation}

Given the exploratory nature of this proof-of-concept study, the dataset was constructed from publicly available high-resolution handwriting and calligraphy videos (1080p, 30 frames per second) sourced from YouTube \cite{DorufaVSArt}. Written permission was obtained from the original content creator for research use. The videos depict only pen motion on paper and do not contain identifiable personal information.

Five videos were selected to cover diverse writing conditions rather than randomly sampled, consistent with the exploratory nature of the study. They included different writing styles (cursive and script), variations in letter shapes and stroke regularity, as well as different pen types (fine and thick tips, black and blue ink). In addition, the paper layouts varied in terms of line spacing and visual structure. 

The videos were parsed at a frequency of 30 frames per second (fps), resulting in a dataset of 13,507 frames. Each frame was manually annotated by three independent annotators using three categories: Pen-Down, Pen-Up, and uninformative frames (e.g., motion blur or ambiguous video transitions). An Inter-Rater Reliability (IRR) reached 89\%, with a Fleiss' Kappa coefficient of 0.78, indicating substantial agreement.

Disagreements were resolved by majority voting for final class assignment, while annotation uncertainty was retained during intermediate processing steps to better account for transition ambiguity. For model training, uninformative frames were removed, resulting in a binary classification task consisting only of Pen-Down and Pen-Up frames. In total, 28.7\% of the frames corresponded to Pen-Up events, reflecting the frame-level class imbalance of the problem.

\subsection{Evaluation Metric}

We report event-level $F_2$ as the primary metric, because the target application is a screening tool in which recall is prioritized over precision: failing to detect a true Pen-Up event is considered more detrimental than producing a small number of false detections.

\subsection{Baseline End-to-end Deep Learning Approach}

As an initial baseline, we evaluated several end-to-end deep learning architectures operating directly on raw video frames, including a 2D CNN (ResNet18), a CNN--LSTM model \cite{donahue_long-term_2016}, and a 3D CNN (R3D-18). These models aim to learn pen-contact states directly from visual and temporal information without explicit kinematic modeling.

Table~\ref{tab:cnn_comparison} summarizes the frame-level performance of these approaches for the Pen-Up class. While the 3D CNN achieved the highest accuracy among the evaluated models, all end-to-end approaches showed limited performance in terms of $F_2$ score, with a maximum of 0.48, indicating limited recall for Pen-Up frames.

\begin{table}[ht]
\centering
\caption{Frame-level performances of end-to-end deep learning models for pen-contact state classification (Pen-Up class)}
\setlength{\tabcolsep}{3pt}
\renewcommand{\arraystretch}{1.0}
\footnotesize
\setlength{\tabcolsep}{9pt}
\begin{tabular}{lcccc}
\toprule
Model & Accuracy & Recall & Precision & $F_2$ score \\
\midrule 
CNN & 0.630 & 0.250 & 0.250 & 0.250 \\
CNN--LSTM & 0.573 & 0.280 & 0.215 & 0.264 \\
3D-CNN & 0.786 & 0.461 & 0.581 & \textbf{0.481} \\
\bottomrule
\end{tabular}
\label{tab:cnn_comparison}
\end{table}

In this limited-data setting, these baseline results suggest that end-to-end image-based models struggle to capture fine-grained pen--paper interaction dynamics robustly across videos with varying visual characteristics, such as differences in writing style, pen appearance, and background structure.

These limitations motivated the adoption of a hybrid approach, in which pen-tip localization and kinematic modeling were explicitly decoupled. By transforming video data into structured trajectories, the proposed pipeline integrates domain-specific information while remaining more interpretable than a fully end-to-end approach.


\subsection{The Four-Stage Pipeline}

Initial experiments using end-to-end deep learning approaches showed limited generalization across videos with different visual characteristics. To address this limitation, we designed a hybrid pipeline that explicitly separated pen-tip localization from pen-state classification, as illustrated in Fig.~\ref{fig:pipeline}. This decoupling allowed the model to operate on structured kinematic signals rather than raw pixel data, with the goal of improving both interpretability and robustness in a low-data regime.


\subsubsection{Stage 1: Pen-Tip Tracking}

The first stage consisted of localizing the pen tip $(u,v)$ in each frame using a fine-tuned YOLOv11m detector \cite{he_performance_2025}. This step is critical, as the entire pipeline relies on the quality and temporal consistency of the extracted trajectory.

For training, one out of every four frames was manually annotated with a bounding box around the pen tip. In LOVO cross-validation, the detector achieved a median localization error of 3.28 px (P95: 7.44 px), with 77.7\% of predictions within 5 px and 99.2\% within 10 px, indicating sufficiently precise localization for downstream kinematic analysis.

\begin{figure}[ht] 
\centering
\includegraphics[width=\columnwidth]{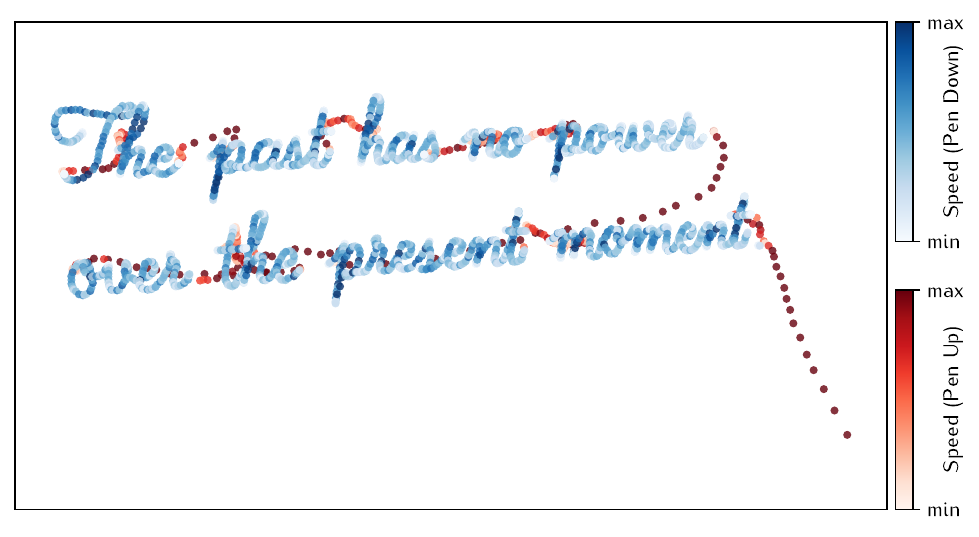}
\caption{
Trajectory of pen movement, with color intensity encoding instantaneous handwriting speed (blue: pen down, red: pen up).}
\label{fig:yolo_and_trajectory}
\end{figure}

After validation, the final YOLOv11m model was trained on the full labeled set using a resolution of $640 \times 640$, batch size of 8, and early stopping (patience of 15, up to 50 epochs). It was then used to infer pen-tip coordinates on the remaining unlabeled frames, yielding a continuous trajectory even during high-lift movements, as illustrated in Fig.~\ref{fig:yolo_and_trajectory}, beyond the proximity range typically captured by digitizing tablets.

\subsubsection{Stage 2: Kinematic Feature Engineering}

From the extracted $(u,v)$ coordinates, we computed a set of 147 kinematic features designed to capture handwriting dynamics at both local and global scales.

\begin{itemize} 

\item \textbf{Local Multi-scale Features:} Computed over sliding windows of varying sizes (from 3 to 16 frames), in order to reflect the temporal variability of Pen-Up events. Features included linearity ratios, directional stability (angular variance), and cumulative sums of acceleration and angular changes to detect sharp state transitions. These features were intended to capture differences between on-surface writing, typically slower and more curved, and in-air movements, which usually tend to be faster and more linear.

\item \textbf{Global Contextual Features:} Computed at the video level to normalize for individual writing styles, these included descriptors such as average baseline slope, mean stroke slant, and velocity Z-scores. These features were intended to reduce sensitivity to inter-subject variability in writing speed and orientation. 

\end{itemize}

This combination of multi-scale local descriptors and global normalization features was introduced to address the limited data setting and the diversity intentionally sought during video selection.


\subsubsection{Stage 3: Model Training and Optimization}

The pen-contact state classification task was formulated as a binary problem (Pen-Down vs. Pen-Up). Several classical machine learning models were evaluated, including Random Forest (RF), Histogram-Based Gradient Boosting (HistGBM) and Light Gradient-Boosting Machine (LightGBM).

Given the limited dataset size, a LOVO cross-validation strategy was adopted. At each iteration, one video was held out for testing, while the remaining videos were used for training. This protocol ensured that the reported performance would reflect the ability of the model to generalize to unseen writing styles and recording conditions.

No separate validation set was used. Instead, hyperparameters were optimized within the cross-validation framework using Optuna \cite{Akiba2019Optuna}, with 100 evaluations per model over predefined search ranges. For this classification stage, the selected configuration maximized the frame-level $F_2$ score.

Each model output a probability $P_{\text{Pen-Up}}$ for each frame. These probabilities were then refined in the next stage to produce temporally coherent predictions.


\subsubsection{Stage 4: Event-Level Optimization}

To align model outputs with clinically meaningful events rather than isolated frame predictions, a post-processing stage was introduced. This stage operated on the predicted probabilities and aimed to reconstruct coherent Pen-Up segments.

The post-processing pipeline relied on three complementary components. 

\begin{itemize} 

\item \textbf{Hysteresis Smoothing:} A dual-threshold scheme was applied to stabilize transitions between Pen-Down and Pen-Up states. This reduced rapid oscillations in the predictions.

\item \textbf{Morphological Filtering:} Successive closing and opening operations were used to merge fragmented segments and remove short spurious detections. 

\item \textbf{Kinematic Alignment:} A physics-inspired heuristic was used to snap predicted segment boundaries to the nearest local velocity minima, under the assumption that pen lifts and touchdowns occur near local speed minima. 

\end{itemize}

To ensure fair evaluation, all post-processing parameters, including hysteresis thresholds and filtering configurations, were optimized within the same LOVO framework as the classification models. This prevented information leakage from the test video and provided a more realistic estimate of generalization performance.

Overall, this stage was designed to convert raw probabilistic predictions into temporally coherent handwriting events that are more interpretable at the clinical level.


\section{Results}

Performance was evaluated at the event level, as the objective was to detect meaningful Pen-Up segments rather than isolated frame predictions. All reported results were obtained under LOVO protocol, ensuring that each test video remained fully unseen during both training and post-processing optimization.

Three temporal tolerances were considered: 5, 10, and 12 frames ($\approx$ 167, 333, and 400 ms at 30 fps). This parameter defines the maximum allowed temporal deviation between predicted and ground-truth events for a correct match. These values cover a practically relevant range for short handwriting pauses, with 12 frames corresponding approximately to the empirical pause duration reported in the literature ($\approx$ 400 ms) \cite{Pascual2023In}. The 5- and 10-frame settings provide stricter matching criteria, whereas 12 frames offers a more clinically aligned reference tolerance.

\subsection{Optimal Thresholding Configuration}
 
The optimal post-processing configurations identified during the LOVO optimization in the Stage 4 grid search for each model are presented in Table~\ref{tab:oracle_params}.  Across all models, hysteresis thresholding proved important for stabilizing predictions and reducing temporal fragmentation.

\begin{table}[ht]
\centering
\caption{Optimal hysteresis thresholds for each model \\ (modal value across folds)}
\label{tab:oracle_params}
\setlength{\tabcolsep}{9pt}
\begin{tabular}{@{}lccc@{}}
\toprule
Model & Low Threshold & High Threshold \\
\midrule
Random Forest    & 0.50 & 0.52 \\
HistGBM   & 0.65 & 0.68 \\
LightGBM  & 0.55 & 0.60 \\
Ridge stacking & 0.50 & 0.52 \\
\bottomrule
\end{tabular}
\end{table}

Interestingly, the kinematic alignment component was not consistently selected in the optimal configurations, indicating that the probabilistic outputs of the classifiers were often already sufficiently aligned with the underlying motion dynamics under the tested tolerances.

\subsection{Event-Level Performance}

Event-level performances for each model are reported in Table~\ref{tab:metrics_tolerance}. Overall, the ensemble-based models (Random Forest, HistGBM, LightGBM) showed the most reliable performance under the LOVO setting.

\begin{table}[htb]
\centering
\caption{Event-level Performances (Pen-Up class) at temporal tolerances of 5, 10 and 12 frames (mean $\pm$ 95\% CI half-width)}
\label{tab:metrics_tolerance}
\renewcommand{\arraystretch}{0.90}
\setlength{\tabcolsep}{3.5pt}
\scriptsize
\begin{tabular}{@{}lcccc@{}}
\toprule
Model & Tol. & Recall & Precision & $F_2$ \\
\midrule
\multirow{3}*{Random Forest}
& 5  & 0.795 $\pm$ 0.145 & 0.526 $\pm$ 0.356 & 0.721 $\pm$ 0.175 \\
& 10 & 0.831 $\pm$ 0.159 & 0.549 $\pm$ 0.356 & 0.753 $\pm$ 0.159 \\
& 12 & 0.838 $\pm$ 0.159 & 0.554 $\pm$ 0.355 & 0.760 $\pm$ 0.164 \\
\midrule
\multirow{3}*{HistGBM}
& 5  & 0.826 $\pm$ 0.070 & 0.553 $\pm$ 0.343 & 0.752 $\pm$ 0.097 \\
& 10 & 0.864 $\pm$ 0.105 & 0.578 $\pm$ 0.339 & 0.786 $\pm$ 0.088 \\
& 12 & 0.868 $\pm$ 0.105 & 0.581 $\pm$ 0.339 & 0.790 $\pm$ 0.091 \\
\midrule
\multirow{3}*{LightGBM}
& 5  & 0.828 $\pm$ 0.079 & 0.564 $\pm$ 0.316 & 0.757 $\pm$ 0.122 \\
& 10 & 0.866 $\pm$ 0.072 & 0.590 $\pm$ 0.311 & 0.792 $\pm$ 0.112 \\
& 12 & 0.880 $\pm$ 0.067 & 0.599 $\pm$ 0.297 & \textbf{0.805 $\pm$ 0.080} \\
\midrule
\multirow{3}*{Ridge stacking}
& 5  & 0.640 $\pm$ 0.256 & 0.723 $\pm$ 0.309 & 0.655 $\pm$ 0.248 \\
& 10 & 0.659 $\pm$ 0.256 & 0.745 $\pm$ 0.289 & 0.674 $\pm$ 0.243 \\
& 12 & 0.661 $\pm$ 0.258 & 0.747 $\pm$ 0.289 & 0.677 $\pm$ 0.246 \\
\bottomrule
\end{tabular}
\end{table}

At a temporal tolerance of 10 frames ($\approx$ 333 ms), gradient boosting methods consistently achieved the best results, with $F_2$ scores above 0.78. Among them, LightGBM obtained the highest $F_2$ score ($0.792$) and the best recall ($0.866$), indicating high sensibility to Pen-Up events. HistGBM achieved slightly lower, but still competitive performances ($F_2 = 0.786$), while Random Forest showed intermediate results ($F_2 = 0.753$). 

At 12 frames ($\approx$ 400 ms), performance increased slightly for all models, with LightGBM reaching the overall best result ($F_2 = 0.805$, Recall = $0.880$). These gains remained modest relative to the 10-frame setting, suggesting that the main performance trends were already established at stricter tolerances.

In contrast, the Ridge stacking approach (with regularization parameter $\alpha = 1$) produced more conservative predictions, favoring precision over recall. Although it achieved the highest precision (up to $0.747$), its lower recall ($0.661$) led to a lower overall $F_2$ score ($0.677$) than the individual ensemble models.

\subsection{Temporal Tolerance Sensitivity Analysis}

The gap between the frame-level and event-level performance indicates that many frame-level prediction errors were due to small temporal misalignments rather than complete detection failures. We report here the results for three tolerances, 5, 10 and 12 frames, although a broader range of values was explored to assess how detection performance varied with the event-matching criterion. 

\begin{figure}[ht]
\centering
\includegraphics[width=\columnwidth]{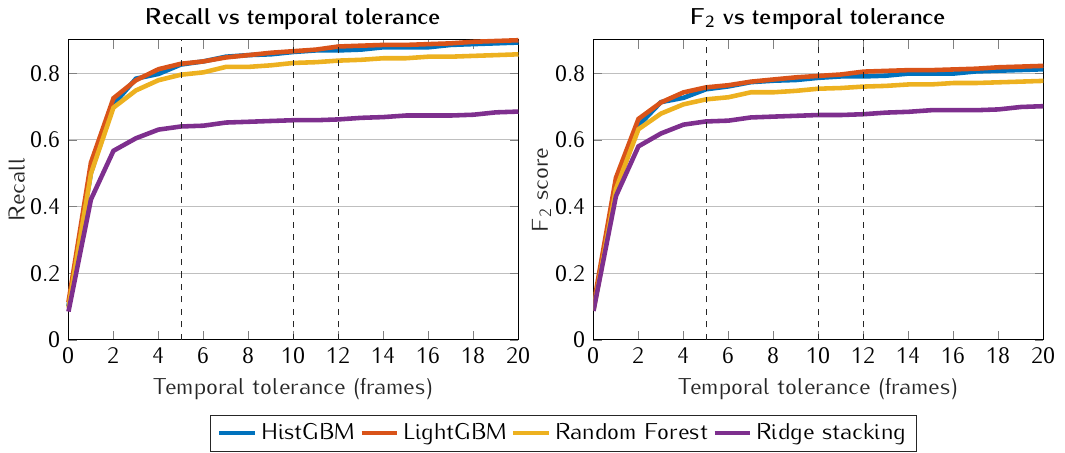}
\caption{Event-level Recall (left) and $F_2$ score (right) as functions of temporal tolerance for the four evaluated models.}
\label{fig:tolerance_curve}
\end{figure}

As illustrated in Fig.~\ref{fig:tolerance_curve}, increasing the temporal tolerance improved Recall and $F_2$ across all models, as the event-matching criterion became less restrictive. The gain was steep at low tolerances and then progressively flattened, suggesting that many residual errors were due to small temporal offsets rather than missed events.

Between 5 ($\approx$ 167 ms) and 10 frames ($\approx$ 333 ms), the average increase remained modest, approximately 0.03 for $F_2$, $0.028$ for Precision, and $0.033$ for Recall. From 10 to 12 frames ($\approx$ 400 ms), the additional gain was smaller, approximately $0.006$ for $F_2$, $0.004$ for Precision, and $0.007$ for Recall on average across models. This pattern indicates that performance was already approaching a plateau around the 10-frame range, even though the best absolute scores were obtained at 12 frames.

This behavior is consistent with slight annotation uncertainty around pen-state transitions and with the temporal smoothing introduced during post-processing. Overall, the results suggest that the proposed method is relatively tolerant to small temporal misalignments, which is desirable in practice while exact frame-level synchronization is difficult to achieve. 

\subsection{Temporal Overlap Analysis via IoU Metrics} 

In addition to event-level detection scores, Intersection over Union (IoU) was used to quantify the temporal overlap between predicted and ground-truth segments \cite{cho_weighted_2024}.

\begin{table}[ht]
\centering
\caption{Mean and Weighted IoU scores \\ at temporal tolerances of 5, 10, and 12 frames}
\label{tab:iou_global}
\renewcommand{\arraystretch}{0.85}
\setlength{\tabcolsep}{9pt}
\begin{tabular}{@{}lccc@{}}
\toprule
Model & Tolerance & Mean IoU & Weighted IoU \\
\midrule
\multirow{3}*{Random Forest}    & 5  & 0.488 & 0.440 \\
      & 10 & 0.473 & 0.415 \\
      & 12 & 0.469 & 0.413 \\
\midrule
\multirow{3}*{HistGBM}   & 5  & 0.484 & 0.452 \\
      & 10 & 0.469 & 0.436 \\
      & 12 & 0.466 & 0.436 \\
\midrule
\multirow{3}*{LightGBM}  & 5  & 0.485 & 0.462 \\
      & 10 & 0.468 & 0.439 \\
      & 12 & 0.461 & 0.424 \\
\midrule
\multirow{3}*{Ridge stacking} & 5  & 0.499 & 0.419 \\
      & 10 & 0.487 & 0.401 \\
      & 12 & 0.485 & 0.400 \\
\bottomrule
\end{tabular}
\end{table}

As tolerance increased, global IoU scores tended to decline slightly, reflecting reduced temporal precision. This trend is shown in Table~\ref{tab:iou_global}, where IoU scores remained relatively stable across models between $0.461$ and $0.499$, with weighted IoU values ranging from $0.400$ to $0.462$.

\subsection{Feature Importance Analysis}

\begin{figure}[ht]
\centering
\includegraphics[width=0.9\columnwidth]{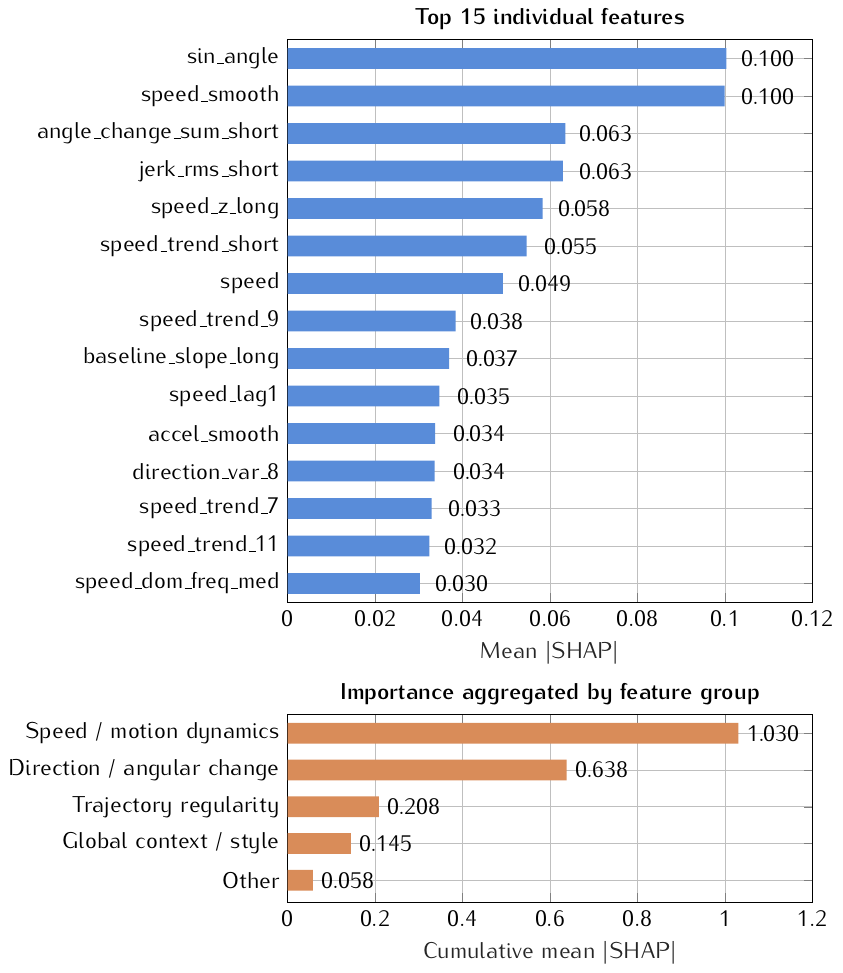}
\caption{Feature importance analysis for the LightGBM model based on mean absolute SHAP values. Top: top 15 individual features. Bottom: cumulative importance aggregated by feature group.}
\label{fig:feature_importance}
\end{figure}

To better understand which kinematic cues supported Pen-Up detection, we examined feature importance for the LightGBM model using mean absolute SHAP values \cite{lundberg2020local}\footnote{\texttt{https://github.com/shap/shap}} (Fig.~\ref{fig:feature_importance}). One advantage of this non-deep learning approach is that the contribution of individual features can be inspected more directly than in end-to-end deep models.

The most influential predictors included smoothed speed, angular variation, jerk, speed trends, and baseline-related context. At the group level, speed- and motion-related features dominated the cumulative importance, followed by direction- and angle-related descriptors. This is consistent with the contrast between on-surface handwriting strokes, typically slower and more curved, and in-air repositioning movements, generally faster and more linear.

The repeated presence of multi-scale features among the top-ranked predictors also indicates that temporal context is important for pen-state inference. Overall, these results support the relevance of the proposed feature engineering strategy and suggest that simple instantaneous thresholding would not be sufficient to capture the variability of handwriting dynamics across videos.


\section{Discussion}

This study explored the feasibility of detecting Pen-Up events from standard top-view videos as a complement to digitizing tablets. As a proof of concept, the goal was not to build a fully optimized system, but to assess whether meaningful kinematic information could be reliably extracted from video in a controlled yet variable setting. The proposed approach follows an event-level prediction framework \cite{guralnik_event_1999}, which is well suited to temporally structured handwriting dynamics.

The results showed that the proposed hybrid pipeline can detect Pen-Up events with reliable performance under a LOVO protocol. In particular, the boosting-based models achieved event-level $F_2$ scores up to $0.805$, indicating effective detection of Pen-Up segments. This is especially relevant in a clinical context, where missing such events could lead to an incomplete characterization of handwriting dynamics.

A key strength of the approach lies in the combination of explicit trajectory extraction and interpretable kinematic features. Unlike end-to-end deep learning baselines, which generalized poorly in this setting, the proposed pipeline relied on simple kinematic properties such as velocity, linearity, and trajectory regularity. This makes the system easier to interpret and likely better suited to low-data regimes, in line with recommendations favoring interpretable models over black-box approaches \cite{Rudin2019Stop}.

Feature importance analysis showed discriminative predictors related to velocity, local motion irregularity, and geometric writing style. The prominence of velocity-based and multi-scale descriptors is consistent with the distinction between Pen-Down and Pen-Up movements, while the global style features such as baseline slope and slant suggest that contextual normalization helps account for variability across heterogeneous conditions.

Although the dataset was limited to five videos, these were intentionally selected to cover diverse writing conditions, including variations in writing styles (cursive and script), pen types, stroke thickness, and paper layouts. In this context, the LOVO evaluation provided a meaningful estimate of generalization within controlled variability.

Beyond predictive performance, the pipeline also offers practical advantages. It can run efficiently on general-purpose CPUs \cite{Schwartz2020Green}, without requiring dedicated GPU resources. In addition, if the original video is discarded after pen-tip trajectories are extracted, data retention can be reduced, which may support privacy-conscious analysis in clinical contexts involving sensitive patient data.

\subsection*{Limitations and Future Work}
Several limitations should nevertheless be acknowledged. First, the dataset remained small, and a larger-scale validation is needed to assess the robustness of the approach more thoroughly. Second, the use of a monocular camera prevented direct estimation of pen height. Pen-Up states were therefore inferred indirectly from kinematic cues, without access to true depth information. Even so, the approach provided trajectory information beyond the proximity range of digitizing tablets. Extending this work to multi-camera setups could enable direct estimation of the third dimension.

Another limitation concerns viewpoint variability. In this study, the camera was positioned in a top-view configuration to simplify the problem and ensure consistent observations. In a real clinical setting, such a setup could be standardized, thereby reducing variability across recordings. More complex scenarios with varying viewpoints would likely require additional data and more advanced modeling strategies.

Finally, the methodological choices reflected the exploratory nature of this work. Hyperparameters were optimized using the Optuna library without a dedicated validation set. This remained acceptable in the present low-data context, where the primary objective was to assess feasibility rather than to report definitive performance benchmarks. Still, 100 hyperparameter evaluations were performed for each model, providing a non-trivial exploration of the search space. 

\section{Conclusion}

In conclusion, by focusing on a simplified Pen-Up/Pen-Down classification task, this work represents an encouraging first step toward machine-learning-based video analysis of handwriting as a complement to digitizing tablets. By enabling trajectory tracking during high-lift movements, the proposed approach helps address an important sensing gap. As a proof of concept, these results provide a foundation for future studies involving larger datasets, controlled acquisition protocols, and potential clinical applications.


\section*{Acknowledgment}

The authors thank the creator of the DorufaVSArt YouTube channel for granting permission to use the handwriting videos included in this study. The authors also thank Romain Sebire for his valuable assistance in the annotation process.


\section*{Data and code availability}

The dataset used in this study cannot be publicly released due to privacy and ethical restrictions. The code developed for this work is not available for public release.


{\small
\bibliographystyle{ieeetr}
\bibliography{bib}
}

\end{document}